\pdfoutput=1

\PassOptionsToPackage{dvipsnames}{xcolor}
\documentclass[11pt]{article}
\usepackage{EMNLP2022}
\usepackage{times}
\usepackage{latexsym}
\usepackage[T1]{fontenc}
\usepackage[utf8]{inputenc}

\usepackage{microtype}
\usepackage{inconsolata}

\usepackage{subcaption}

\usepackage{graphicx}
\usepackage{amsbsy}
\usepackage{vcell}
\usepackage{multirow}
\usepackage{amsmath}
\usepackage[ruled,vlined]{algorithm2e}
\SetInd{0.2em}{0.3em}
\SetArgSty{textup}

\usepackage{booktabs}

\usepackage[normalem]{ulem}
\useunder{\uline}{\ul}{}
\usepackage{enumitem}

\title{Human-Machine Collaboration Approaches to Build a Dialogue Dataset for Hate Speech Countering}

\author{Helena Bonaldi$^{1,3}$, Sara Dellantonio$^{2,3}$, Serra Sinem Tekiro\u{g}lu$^{3}$, Marco Guerini$^{3}$,  \\
  $^1$University of Trento, Italy \\
  $^2$Free University of Bozen-Bolzano, Italy\\
 $^3$Fondazione Bruno Kessler, Via Sommarive 18, Povo, Trento, Italy \\
  \texttt{hbonaldi@fbk.eu}, \texttt{sdellantonio@fbk.eu}, \texttt{tekiroglu@fbk.eu}, \texttt{guerini@fbk.eu}}

\date{}

\begin{document}
\maketitle

\begin{abstract}
Fighting online hate speech is a challenge that is usually addressed using Natural Language Processing via automatic detection and removal of hate content. Besides this approach, counter narratives have emerged as an effective tool employed by NGOs to respond to online hate on social media platforms. For this reason, Natural Language Generation is currently being studied as a way to automatize counter narrative writing. However, the existing resources necessary to train NLG models are limited to 2-turn interactions (a hate speech and a counter narrative as response), while in real life, interactions can consist of multiple turns. In this paper, we present a hybrid approach for dialogical data collection, which combines the intervention of human expert annotators over machine generated dialogues obtained using 19 different configurations. The result of this work is DIALOCONAN, the first dataset comprising over 3000 fictitious multi-turn dialogues between a hater and an NGO operator, covering 6 targets of hate. 
\end{abstract}

\section{Introduction}

\begin{figure}[t]
\centering
{\includegraphics[width=1\columnwidth]{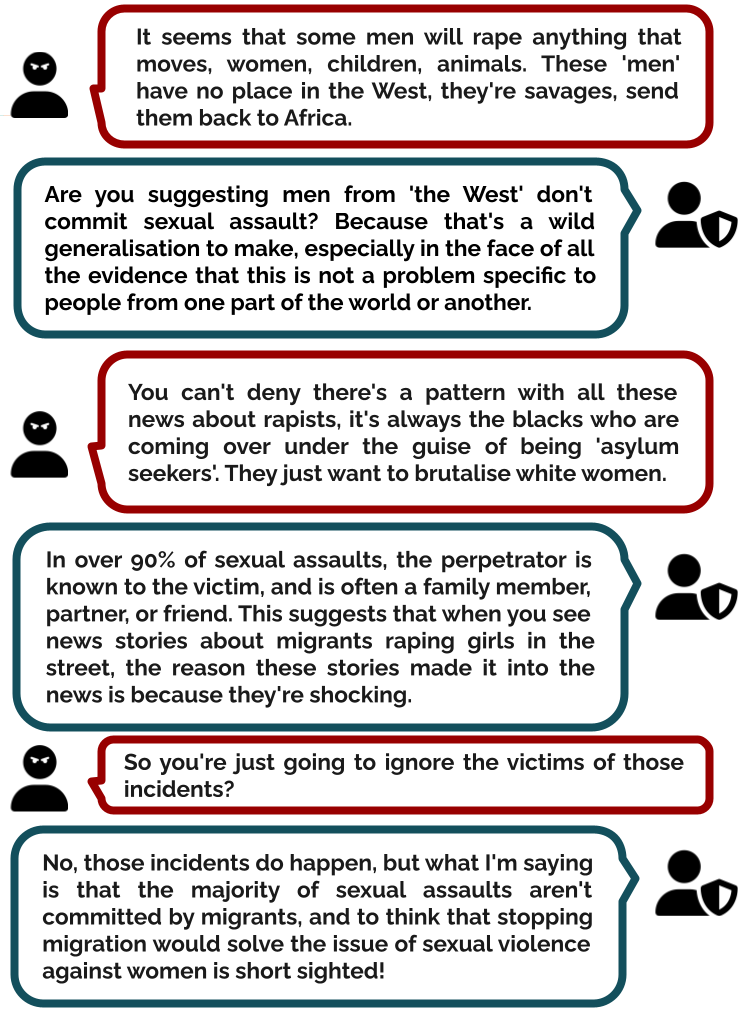}}
\caption{Exemplar dialogue between a hater and an NGO operator.}
\label{fig:dialogue_ex}
\end{figure}

While hate towards vulnerable groups or individuals is not a new phenomenon, the upsurge of hate speech and its proliferation is relatively recent and it is enabled by the fast spread of information in online platforms. The rise in hate speech online can even provoke violent actions offline. Consequently, fighting online Hate Speech (HS) has become a vitally important ``job for everyone''\footnote{``Hatred is a danger to everyone – and so fighting it must be a job for everyone.'' \textit{António Guterres, United Nations Secretary-General, 2021}} especially for the NLP researchers. 
The contrast to HS and haters on social media platforms is usually carried on via user suspension, content removal or shadow banning, which can be mapped to a classification task in NLP terms. However, AI and NLP can play even a more crucial role that is not limited to classification. In fact, recently, NLG models have started to be proposed as an effective tool to counter HS by providing relevant responses. In particular, the idea is to imitate the operators of Non-Governmental Organizations (NGO) that are actually intervening in online discussions by replying to hateful content using so-called Counter Narratives (CN), defined by \citet{schieb2016governing} as ``communicative actions aimed at refuting hate speech through thoughtful and cogent reasons, and true and fact-bound arguments''. Through automatically generating CNs, it is possible to aid NGO operators in their day-to-day manual activities, and therefore to partially countervail the sheer amount of hateful content posted online \cite{chung2021empowering}. 

Despite the invaluable attempts to create HS/CN datasets and systems \citep{mathew2019thou,qian-etal-2019-benchmark,conan-2019,fanton2021human}, up to now only datasets containing 2-turn interactions have been proposed (i.\ e.\ a hate speech and a responding counter narrative), while in real scenarios, such as on social media platforms, multi-turn dialogues are the norm. In Figure \ref{fig:dialogue_ex} an example of such dialogues is provided\footnote{This paper includes examples of hateful content, which may be upsetting for the readers. However, they do not represent the views of the authors.}. Therefore, multi-turn dialogue datasets are necessary for training models that can better handle online hate phenomenon. 

Still, obtaining expert-written quality data to train such models on is not trivial. To ameliorate this problem, a recently proposed approach is the use of \textit{hybrid} data collection strategies where a human and a machine collaborate to build data starting from a seed dataset of expert based examples \citep{fanton2021human}. In this paper we follow this line of research and investigate novel strategies and algorithms that are specifically designed for multi-turn dialogues collection.

In particular, we test 19 different \textit{hybrid} strategies obtaining a novel dataset of more than 3K dialogical interactions between two interlocutors, one acting as the hater and the other as the NGO operator, for a total of more than 16K turns. We call this dataset DIALOCONAN (DIALOgical COunter-NArratives collectioN). This is the first and most comprehensive multi-target dataset that addresses expert-based counter narrative generation in fully dialogical scenarios, and it can be downloaded at the following link: \url{https://github.com/marcoguerini/CONAN}.

\section{Related Work}

In this work, we consider four main research areas as relevant: in particular (i) available datasets for hate speech detection, (ii) available datasets for CN generation, (iii) CN generation approaches, and (iv) hybrid data collection methodologies.

\paragraph{Hate detection.} Many benchmarks for automatic HS detection are currently available  \citep{mathew2021hatexplain, cao2020deephate, kumar2018benchmarking,hosseinmardi2015detection, waseem2016you, burnap2016us}. Regarding the systems built on top of these benchmarks, we refer the readers to the surveys by \citet{poletto2020resources, schmidt2017survey, nunes2018survey} for detailed reviews. Other reviews include the analysis of ethical implications \citep{kiritchenko2020confronting} and of problems such as bias replication \citep{10.1007/978-3-319-67256-4_32,  davidson2019racial, vidgen2020directions, sap2019risk, tsvetkov2020demoting}. 

\paragraph{CN data collection.} Since CNs have been shown to be effective in reducing linguistic violence \citep{benesch2014countering, gagliardone2015countering, schieb2016governing, silverman2016impact, mathew2019thou} and in changing the viewpoints of bystanders \citep{allison2016cyber,anderson2014combating}, they are beginning to be collected as training data for supervised NLG models. The investigated approaches for data collection can be listed as crawling \citep{mathew2018analyzing, mathew2019thou, yu-etal-2022-hate}, crowdsourcing \citep{qian-etal-2019-benchmark},  nichesourcing  \citep{conan-2019} and hybrid approaches  \citep{tekiroglu-etal-2020-generating, fanton2021human}. 

The most relevant datasets for our work are (i) \citet{fanton2021human} in terms of quality and target diversity, even if it only includes HS/CN pairs, and (ii) \citet{qian-etal-2019-benchmark} that hints at the issue of multi-turn dialogues. However, in the latter the CN is only the last turn of a forum-style dialogue among more than 2 interlocutors, rather than a HS/CN multi-turn dialogue between two opposing actors. 

\paragraph{CN generation.} Neural approaches to generate CNs have started to be studied along with available datasets \citep{fanton2021human, tekiroglu-etal-2020-generating, qian-etal-2019-benchmark}. \citet{tekiroglu2022using} present a thorough comparison of several pre-trained LMs for this task. \citet{zhu2021generate} propose an entirely automated 2 stage pipeline where several CN candidates are generated and then filtered. Other lines of work include CN generation for under-resourced languages \citep{chung2020italian}, or the generation of knowledge-bound CNs, to avoid hallucination phenomena \citep{chung-etal-2021-towards}. Finally, \citet{ashida2022towards} studied CN generation with LLMs, using few-shots prompting.

\paragraph{Hybrid models for data collection.} A recently emerged data collection methodology is based on \textit{hybrid} models, where humans and machines work together to collect better quality data in a more efficient way. \citet{GAN_hum_loop} propose using model output to guide humans in the writing of adversarial examples for question-answering systems. \citet{dinan2019build} and \citet{vidgen2020learning} perform a data collection for offensive language detection with repeated model-human interactions where the classifier output drives annotators in example creation at each round. A more recent study proposes a \textit{hybrid} approach where an LM is trained to generate HS/CN pairs that are validated and post-edited by annotators \citep{tekiroglu-etal-2020-generating}. \citet{fanton2021human} further expand this approach by making it iterative using several LM configurations.

\section{Methodology} \label{methodology}

Data collection can be very difficult and time consuming when high quality data from experts are necessary. Given that we need to collect whole HS/CN dialogues and not just pairs, the problem is even harder. Moreover, scraping  NGO operators' real interactions is not a viable solution, considering that this data can be used for account ``doxing''. In fact, malicious users could reverse-search the text included in a dataset to identify the operators' accounts. This would undermine their work, since they usually operate undercover, and would expose them to possible attacks. 

Therefore, we decided to resort to \textit{hybrid} approaches and run 3 different data collection sessions based on the aspects of the dialogue augmentation we want to address (either the structure, in terms of turns order, or the wording of the turns). 

In total we tested 19 different dialogue collection strategies. All the strategies are inserted in an author-reviewer pipeline as described by \citet{tekiroglu-etal-2020-generating}, where the \textit{author} is a single dialogue creation strategy at a time, and the \textit{reviewer} is represented by a team of trained annotators, who are tasked with post-editing the dialogues generated by the given author strategy.

\paragraph{Author - Configurations.} Each of the 3 data collection sessions we perform has different input data and author tasks, in particular:

\begin{itemize}[leftmargin=*]
    \item \textbf{Session 1: same wording, new dialogue structure.} 7 strategies based on concatenating pre-existing material (HS/CN pairs) to obtain new dialogues. 
    \item \textbf{Session 2: new wording, same dialogue structure.} 6 strategies to modify the wording of pre-existing dialogues via paraphrasing.
    \item \textbf{Session 3: new wording, new dialogue structure.} 6 strategies using generative Language Models (LMs) for complete dialogue generation.
\end{itemize}

\paragraph{Author - Seed datasets.} Since each author configuration needs some textual input, we employ (i) a dataset, created ad hoc, consisting of 222 fictitious dialogues and (ii) HS/CN pairs coming from the dataset presented in \citet{fanton2021human}. 

The ad hoc fictitious dialogues (DIALO$_{gold}$ henceforth) are written by two expert NGO operators, who have been working for over 10 years in writing CNs on social media platforms. They were asked to write dialogues between a hypothetical hater and an NGO operator, following their real expertise in the task.  
The dialogues can have 4, 6, or 8 turns (these  are typical lengths according to their experience) and cover the following 6 targets of hate, defined beforehand: \texttt{LGBT+}, \texttt{MIGRANTS}, \texttt{MUSLIMS}, \texttt{JEWS}, \texttt{POC} and \texttt{WOMEN}. 

Given the small size of DIALO$_{gold}$, we also use part of the dataset presented in \citet{fanton2021human} as an additional resource. This dataset consists of 5000 HS/CN pairs covering, among others, the 6 targets of hate present in DIALO$_{gold}$. Therefore, we extracted the pairs labeled with these 6 targets so that the two resources can be `aligned' by topic, and we named it PAIRS$_{gold}$, since also this dataset was created with the help of expert NGO operators.

\paragraph{Reviewers - Training.} For post-editing the output of the various author configurations, three annotators were recruited from a pool of internship students. They have been extensively trained using the methodology of \citet{fanton2021human}, in order to become ``experts'' on HS/CN post-editing. In particular, we first explained the aim of the task. Then, they had to read NGO guidelines and documentation on CN writing\footnote{See \url{https://getthetrollsout.org/stoppinghate} as a reference.}, together with all the dialogues present in DIALO$_{gold}$, which were provided as examples of the material they would have to work with. 
We detailed the methodology, explaining that the main focus was to make the dialogues natural, with the minimum intervention possible and keeping the seed dataset as a reference for naturalness. General instructions about the post-editing procedure were also provided, pointing out that for each session specific guidelines would have been given. 

\paragraph{Reviewers - Mitigation procedure.} Finally, we also implemented a mitigation procedure similar to the one presented by \citet{vidgen-etal-2019-challenges}. This procedure is implemented to safeguard the annotators' well-being while working with abusive content and it includes: (i) explaining to the annotators the pro-social nature of the research and the purpose of their post-editing activity, (ii) advising the annotators to work few hours per day and to take regular breaks (iii) having weekly meetings to let possible problems or distress emerge. 

\paragraph{Data collection procedure.} For each session we applied the following procedure: (i) generate dialogue candidates  according to session specific strategies, (ii) adapt the annotation guidelines to the specific session, (iii) let the annotators practice the task on a small ``training'' set of dialogue candidates, and (iv) update the guidelines with respect to their feedback. Lastly, (iv) annotators complete the post-editing on the remaining dialogues following the updated guidelines (the order of the dialogues was randomized to avoid comparison or primacy/recency effects over session strategies).

\section{Metrics} 
We use several metrics to assess the performance of each strategy. These metrics are aimed to assess either the \textit{efficiency} of the procedure or the \textit{quality} of the obtained data. \\

\noindent\textbf{HTER} is an efficiency metric used to measure the post-editing effort of the annotator, and it is usually employed for sentence level translations \cite{specia2010estimating}. A value above 0.4 is generally used to account for low quality outputs, where rewriting from scratch is on par with correcting it \cite{turchi2013coping}. 

\noindent\textbf{Turn deletion} is the percentage of turns that are discarded by the reviewers since their quality is too low and/or they do not fit in the current dialogue structure. The more content needs to be deleted, the less efficient the procedure is.

\noindent\textbf{Turns swap} is the percentage of turns that are moved by the reviewers from the original position they were in, to another position in the final edited dialogue. Usually turns of this kind have a good quality but they do not fit the current position.

\noindent\textbf{Novelty} is utilized to check the quality of a generated dialogue by measuring its lexical difference with respect to a reference set of dialogues, and it is grounded on Jaccard similarity \citep{dziri2018augmenting,wang2018sentigan}.

\noindent\textbf{Repetition Rate} (RR) measures the language diversity within a corpus using the rate of non-singleton ngram types \citep{cettolo2014repetition,bertoldi2013cache}. It is used in our experiments to evaluate each strategy in terms of its ability to provide diverse and varied examples.
 
\section{Session 1: Dialogue structure}
In Session 1 we started from the HS/CN pairs in PAIRS$_{gold}$ and concatenated them in order to produce dialogue candidates with different structures.

\subsection{Author Strategies} We employ 7 strategies to connect HS/CN pairs from PAIRS$_{gold}$ to create 4, 6, and 8 turns examples (consistently with the DIALO$_{gold}$ characteristics). 

During the concatenation, each pair is used only once in a dialogue. The connection strategies are: random concatenation (1 strategy), similarity concatenation (4 strategies), and keyword matching concatenation (2 strategies).  
In order to obtain a balanced dataset, for each strategy, each target, and each dialogue length combination we created 10 connected dialogues. In-detail descriptions of the 7 concatenation strategies we utilized are as follows: 

\paragraph{Random connection.}
For the random connection (\textbf{RND}), the selected pairs for each target are randomly concatenated to form dialogues. This strategy represents a baseline to which we compare against while analysing the other strategies. 

\begin{table*}[ht!]
\begin{center}
\small
\begin{tabular}{l|rrr|rrrr}
\hline
& \multicolumn{3}{c|}{\textbf{Efficiency}} & \multicolumn{4}{c}{\textbf{Quality}} \\ \hline
\multicolumn{1}{r|}{\textbf{}} & \multicolumn{1}{c}{\textbf{del turns}} & \multicolumn{1}{c}{\textbf{HTER}} & \multicolumn{1}{c|}{\textbf{swap}} & \multicolumn{1}{c}{\textbf{RR$_{gen}$}} & \multicolumn{1}{c}{\textbf{RR$_{ed}$}} & \multicolumn{1}{c}{\textbf{NOV$_{g\text{-}g}$}} & \multicolumn{1}{c}{\textbf{NOV$_{g\text{-}e}$}} \\
RND                                    & 12.222                                                  & \textbf{0.141}                                              & 20.926                                              & \textbf{\underline{4.286}}                                                    & \textbf{\underline{4.482}}                                                   & 0.820                                                     & 0.818                                                     \\
J-SIM$_{HS\text{-}HS}$                   & 14.259                                                  & 0.193                                              & \textbf{15.185}                                              & 9.353                                                    & 5.964                                                   & 0.827                                                     & 0.823                                                     \\
C-SIM$_{HS\text{-}HS}$                  & 10.370                                                  & 0.186                                              & 15.926                                              & 9.450                                                    & 5.215                                                   & 0.824                                                     & 0.820                                                     \\
KW$_{HS\text{-}HS}$                      & 21.111                                                  & 0.283                                              & \underline{\textbf{14.444}}                                              & 15.774                                                   & 4.710                                                   & \underline{\textbf{0.828}}                                                     & 0.823                                                     \\
J-SIM$_{CN\text{-}HS}$                   & 10.741                                                  & 0.145                                              & 23.333                                              & 7.454                                                    & 6.204                                                   & 0.826                                                     & \underline{\textbf{0.824}}                                                     \\
C-SIM$_{CN\text{-}HS}$                    & \textbf{8.889}                                                   & \underline{\textbf{0.134}}                                              & 18.704                                              & \textbf{7.128}                                                    & \textbf{5.087}                                                   & 0.820                                                     & 0.818                                                     \\
KW$_{CN\text{-}HS}$                       & \underline{\textbf{8.197}}                                                   & 0.152                                              & 15.222                                              & 11.710                                                   & 9.035                                                   & \underline{\textbf{0.828}}                                                     & \underline{\textbf{0.824}}                                                    
\end{tabular}
\caption{Results for the first session. J-SIM and C-SIM are the connections via Jaccard and cosine similarity, respectively. RR$_{gen}$ and RR$_{ed}$ are respectively the RR of the data before and after post-editing, while NOV$_{g\text{-}g}$ and NOV$_{g\text{-}e}$ are the novelty of the data before and after post-editing with respect to DIALO$_{gold}$.}
\label{tab:session1_result}
\end{center}
\end{table*}

\paragraph{Similarity connection.}
To connect pairs depending on to their similarity, we utilize (i) the Jaccard similarity and (ii) the cosine similarity\footnote{Cosine Similarity is computed on their embeddings obtained with \texttt{mpnet-base}. The Sentence Transformer library (\url{https://www.sbert.net/}) has been employed.}. Both for the Jaccard and cosine similarity, we perform pair matching via two approaches to form the $HS_i, CN_i, HS_{i+1}, CN_{i+1}$ concatenation:

\begin{enumerate}
    \item \textbf{SIM$_{HS\text{-}HS}$} = the similarity between $HS_i$ and $HS_{i+1}$;
    \item \textbf{SIM$_{CN\text{-}HS}$} = the similarity between $CN_i$ and $HS_{i+1}$;
\end{enumerate}

For each pair, we randomly select 1 among the 10 most similar pairs according to the chosen similarity (either Jaccard or cosine) and concatenation elements (either
HS-HS or CN-HS). The procedure is repeated until the desired number of turns for each dialogue is reached. 

\paragraph{Keywords connection.}
We employ the YAKE keyword extractor \citep{campos2020yake} to extract two keywords from each HS and CN of PAIRS$_{gold}$ and perform a concatenation similar to the previous strategies. We connect $HS_i, CN_i$ and $HS_{i+1}, CN_{i+1}$ according to the following criteria:

\begin{enumerate}
    \item \textbf{KW$_{HS\text{-}HS}$} = if $HS_i$ and $HS_{i+1}$ share two keywords;
    \item \textbf{KW$_{CN\text{-}HS}$} = if $CN_i$ and $HS_{i+1}$ share two keywords;
\end{enumerate}

We decided on a 2-keywords match since according to our preliminary manual analysis we found that the first keyword is often target-related; by considering two keywords we aim to include also a topic-related keyword.

As a final note, we should highlight that the two groups of connection strategies (HS-HS and CN-HS) represent either (i) a \textit{global} semantic coherence across turns (all HS being similar) or (ii) a \textit{local} semantic coherence (only CN-HS of adjacent turns being similar) both for SIM and KW.
By using a \textit{global} semantic coherence via HS-HS matching  we attempted to simulate the attitude of the hater which is convinced of their own ideas and do not accept any external input, while with the \textit{local} connections, we aimed to recreate a ``linguistic alignment'' phenomenon \cite{doyle2016investigating}.
Details on the matching procedures and the description of the algorithms for SIM and KW we employed are reported in Appendix \ref{algo_concatenation}.

\subsection{Reviewing phase and guidelines}
In order to obtain natural dialogues, the annotators in this session received specific post-editing instructions:

\begin{enumerate}[leftmargin=*]
    \item Since CNs are gold, it is strongly suggested to post-edit only the HS$_{i+1}$ to ``align'' it with the CN$_{i}$ belonging to the previous turn.
    \item If a pair is in an unnatural position of the dialogue it should be moved to a better position.
    \item If a pair is not fitting with the flow of the dialogue and cannot be moved elsewhere, it should be deleted.
    \item If the whole dialogue makes no sense, or is too difficult to fix, it should be deleted.
\end{enumerate}

A characteristic example of the post-editing done in Session 1 is shown in Table \ref{tab:session1_example1} in Appendix \ref{reviewing_examples}.

\begin{table*}[ht!]
\begin{center}
\small
\begin{tabular}{l|c|rrrrrr} 
\hline
          & \multicolumn{1}{c|}{\textbf{Efficiency}}    & \multicolumn{6}{c}{\textbf{Quality}}         \\ 
\hline
\multicolumn{1}{r|}{} & \multicolumn{1}{c|}{\textbf{HTER}} & \textbf{RR$_{gen}$} & \textbf{RR$_{ed}$} & \textbf{NOV$_{g\text{-}g}$} & \textbf{NOV$_{g\text{-}e}$} & \textbf{NOV$_{mt\text{-}g}$} & \multicolumn{1}{c}{\textbf{NOV$_{mt\text{-}e}$}}  \\
Neutral$_1$                           & 0.355                                               & 4.019                                                    & 3.684                                                   & 0.749                                                    & 0.770                                                   & 0.258                                                  & 0.450                                                  \\
Neutral$_2$                           & 0.398                                               & \textbf{3.943}                                                    & \underline{\textbf{3.275}}                                                   & \underline{\textbf{0.775}}                                                    & \underline{\textbf{0.774}}                                                   & \underline{\textbf{0.470}}                                                  & \underline{\textbf{0.472}}                                                  \\
Style$_{tw}$                            & 0.355                                               & \underline{\textbf{3.836}}                                                    & \textbf{3.396}                                                   & 0.756                                                    & 0.773                                                   & 0.327                                                  & 0.465                                                  \\
Style$_{dialo}$                         & \textbf{0.348}                                               & 5.452                                                    & 4.253                                                   & 0.743                                                    & 0.765                                                   & 0.388                                                  & 0.465                                                  \\
Style$_{formal}$                        & \underline{\textbf{0.332}}                                               & 4.512                                                    & 3.710                                                   & 0.751                                                    & 0.764                                                   & 0.359                                                  & 0.450                                                  \\
Style$_{casual}$                        & 0.369                                               & 4.346                                                    & 4.118                                                   & \textbf{0.763}                                                    & \underline{\textbf{0.774}}                                                   & \textbf{0.416}                                                  & \textbf{0.468}                                                 
\end{tabular}
\caption{Results for the second session. The showed paraphrasers are, from top to bottom: the Protaugment and Style paraphraser with basic, Twitter and Switchboard style, and the Style former paraphrasers with formal and casual style. NOV$_{mt\text{-}g}$ and NOV$_{mt\text{-}e}$ are  the novelty of the generated and post-edited data with respect to the dialogues resulting from Session 1.}
\label{tab:para_res}
\end{center}
\end{table*}

\subsection{Results}

Results of this session in terms of \textit{efficiency} and \textit{quality} are reported in Table \ref{tab:session1_result}. In general, we observe that strategies using any HS-HS connection are less efficient, having higher HTER scores as compared to the CN-HS ones. HS-HS connections also have a high rate of deleted turns, in particular KW$_{HS\text{-}HS}$ and J-SIM$_{HS\text{-}HS}$. The KW$_{HS\text{-}HS}$ strategy is even more inefficient than the random connection baseline (it reaches the highest number of deleted turns and the highest HTER), and it is the most repetitive before post-editing, as showed by the RR$_{gen}$. These results are also confirmed by the annotators' feedback, who noted the presence of dialogues which were particularly difficult to edit since they contained the same HS repeated multiple times (see example in Table \ref{tab:session1_example3}, Appendix \ref{reviewing_examples}). A \textit{posteriori} analysis showed that these dialogues were mainly obtained through the KW$_{HS\text{-}HS}$ connection. Moreover, each HS-HS connection strategy achieves a higher RR$_{gen}$ score than its CN-HS counterpart, showing that connecting through a \textit{global} similarity generates a higher overall repetitiveness than using a \textit{local} similarity. The particular high scores reached by the RR$_{gen}$ of both the keywords connection strategies can be explained by the procedure employed for connection: for keywords we performed an exact matching, whereas with the cosine or Jaccard similarity, the connection was selected from the 10 most similar candidates.

After post-editing, all the strategies achieve a lower RR, between 4.5 and 9, indicating a more diversified content. The novelty is calculated against DIALO$_{gold}$: the scores are similar for all the strategies, and they are hardly affected by the post-editing, showing that each strategy managed to add a consistent novelty to the already present gold data. Finally, it is worth noting that the strategies employing HS-HS connections have less turn swaps than C-SIM$_{CN\text{-}HS}$ and J-SIM$_{CN\text{-}HS}$. The most probable explanation is that CN-HS strategies require less deletion, but this comes at the cost of more turn swaps.

\section{Session 2: Dialogue Wording}
In the second session we focused on strategies aiming to obtain a new wording, given a structured dialogue. In particular, we tested 6 paraphrasing approaches on DIALO${gold}$ and on a part of the dialogues resulting from the first session. In this session, our overall aim is to obtain novel and diverse responses to hate. Therefore, we chose to paraphrase only the CNs belonging to a subset of the data collected in Session 1, while keeping the corresponding HS as it is.

\subsection{Author Strategies}
We carried out two exploratory studies to test different paraphrasing configurations and we selected the 6 most promising ones, as described in Appendix \ref{exploratory_paraphrases}. We use both paraphrasers with no specific style and with style transfer in order to attain a diverse data collection. The selection has been performed by assessing the aspects of dialogue wording that we deem the most relevant for our scenario.

\paragraph{Basic paraphrasing.} We use 2 paraphrasing tools as a `baseline' where we do not impose any specific style to the paraphrases: the Protaugment paraphraser \citep{dopierre2021protaugment} and the Style paraphraser \citep{krishna2020reformulating} with basic style. 
    
\paragraph{Style paraphrasing.} This group includes 4 strategies in which we aimed to generate paraphrases with specific styles, in order to enhance the diversity of our data collection. Specifically, we focused on a style similar to that present in social media or in dialogues (Style paraphraser \citep{krishna2020reformulating} with Twitter and Switchboard style), and formal or casual (Style former paraphraser\footnote{ \url{https://github.com/PrithivirajDamodaran/Styleformer}} with casual and formal style).\\
  
For each CN, 3 different paraphrases are generated using the same paraphrasing strategy.

\subsection{Reviewing phase and guidelines}
In order to obtain more natural examples, the post-editing instructions given to the annotators are adapted accordingly, emphasizing the significance of novel wording.

\begin{enumerate}[leftmargin=*]
    \item The annotator should keep the gold HS as it is, while post-editing the most promising among the 3 CN paraphrasis suggestions, i.\ e.\ the one introducing the least errors and the most different one from the original. 
    \item Turn swap in this case is not allowed, since turns order was already validated in these dialogues and paraphrasing would not affect it. 
    \item For the same reason, turn and dialogue deletion are not allowed.
\end{enumerate}

An example of a typical intervention of the annotators in Session 2 is shown in Table \ref{tab:session2_example1} in Appendix \ref{reviewing_examples}.

\begin{table*}[htbp!]
\begin{center}
\small
\begin{tabular}{l|rrr|rrrrrr} 
\hline
          & \multicolumn{3}{c|}{\textbf{Efficiency}}            & \multicolumn{6}{c}{\textbf{Quality}}                  \\ 
\hline
\multicolumn{1}{c|}{} & \multicolumn{1}{l}{\textbf{del turns}} & \multicolumn{1}{l}{\textbf{HTER}} & \multicolumn{1}{l|}{\textbf{swap}} & \multicolumn{1}{c}{\textbf{RR$_{gen}$}} & \multicolumn{1}{c}{\textbf{RR$_{ed}$}} & \multicolumn{1}{c}{\textbf{NOV$_{t\text{-}g}$}} & \multicolumn{1}{c}{\textbf{NOV$_{t\text{-}e}$}} & \multicolumn{1}{c}{\textbf{NOV$_{g\text{-}g}$}} & \multicolumn{1}{c}{\textbf{NOV$_{g\text{-}e}$}}  \\
DGPT$_b$                              & 50.179                                                  & 0.678                                              & 8.214                                               & \underline{\textbf{7.815}}                                                    & \underline{\textbf{3.976}}                                                   & \textbf{0.793}                                                     & 0.804                                                    & 0.787                                                    & 0.793                                                    \\
DGPT$_{mt}$                             & \underline{\textbf{16.786}}                                                  & 0.408                                              & 10.714                                              & \textbf{8.587}                                                    & 6.110                                                   & 0.757                                                     & 0.759                                                    & 0.798                                                    & 0.799                                                    \\
T5$_{b\text{-}1m}$                             & 76.875                                                  & 0.655                                              & 0                                                   & 16.672                                                   & \textbf{5.651}                                                   & 0.789                                                     & \underline{\textbf{0.817}}                                                    & 0.783                                                    & \underline{\textbf{0.804}}                                                    \\
T5$_{mt\text{-}1m}$                            & \textbf{34.375}                                                  & \underline{\textbf{0.362}}                                              & 0                                                   & 10.605                                                   & 6.950                                                   & 0.756                                                     & 0.756                                                    & \underline{\textbf{0.802}}                                                    & \textbf{0.803}                                                    \\
T5$_{b\text{-}2m}$                             & 85.000                                                  & 0.603                                              & 0                                                   & 20.658                                                   & 5.764                                                   & \underline{\textbf{0.804}}                                                     & \textbf{0.805}                                                    & 0.793                                                    & 0.794                                                    \\
T5$_{mt\text{-}2m}$                            & 38.929                                                  & \textbf{0.376}                                              & 0                                                   & 10.756                                                   & 7.678                                                   & 0.756                                                     & 0.756                                                    & \textbf{0.799}                                                    & \textbf{0.803}                                                   
\end{tabular}
\caption{Results for the third session: the baseline models are signaled by the subscript $_b$, while the models trained on both DIALO$_{gold}$ and the dialgues resulting from Session 1 have the subscript $_{mt}$. NOV$_{t\text{-}g}$ and NOV$_{t\text{-}e}$ are respectively the novelty scores of the generated and post-edited data with respect to each model's training data.}
\label{tab:session3_results}
\end{center}
\end{table*}

\subsection{Results}

We report the results in terms of \textit{efficiency} and \textit{quality} in Table \ref{tab:para_res}.
All the paraphrasers employed reach similar HTER scores, which are below the 0.4 threshold, but higher than Session 1 results. Regarding the quality, generated paraphrases are highly novel with respect to the dialogues present in DIALO$_{gold}$, but not as high if compared to the dialogues resulting from the connection of the gold pairs in Session 1. In addition, the annotators' intervention enhances the novelty of the generated paraphrases in almost all the cases, and reduces the RR for all the paraphrasers, with lower scores than in the first session (3,739 vs. 5,814 on average).

To sum up, we conclude that it is better to concatenate PAIRS$_{gold}$ if we have a high number of pairs available, while paraphrasing is a viable solution if there is no pairs availability, since it implies a higher HTER and it is not justified by higher novelty. 

\section{Session 3: Generation}
In this session we follow the overall configuration presented in \citet{fanton2021human}, where the author is an LM fine-tuned on the DIALO$_{gold}$ together with the dialogues resulting from Session 1\footnote{We did not include the dialogues resulting from Session 2 since they would have added little novelty. In particular, Session 2 CNs are paraphrases of those present in Session 1, and the HS of the dialogues in the two sessions are identical.}. 

\subsection{Author Strategies}
We tested the following configurations: 

\paragraph{DialoGPT.} An autoregressive model specific for dialogue generation \citep{zhang2020dialogpt}. We choose DialoGPT since it is proven to be effective in CN generation as well \citep{tekiroglu2022using};
    
\paragraph{T5$_{2m}$.} Two T5 \citep{raffel2020exploring} models conversing with each other: one fine-tuned to produce only HS and one to produce CNs. This configuration allows to completely decouple CN production from HS production.

\paragraph{T5$_{1m}$.} One T5 model able to produce both HS and CN. We test it as a comparison to the two T5 models conversing with each other.\\

For each configuration, we test a baseline model, fine-tuned on DIALO$_{gold}$ only, and a model fine-tuned on both DIALO$_{gold}$ and the post-edited dialogues resulting from Session 1. For each model we employed the Top-$p$ decoding mechanism \citep{holtzman2019curious} with $p=0.9$\footnote{Training details are reported in Appendix \ref{training_details}.}. In all cases, we split the employed dataset into training, development, and test sets with a ratio of 8:1:1. For the generation phase, we use as a prompt the initial HS of the test set dialogues. Then, we generate a single turn at a time by feeding the model with the context generated so far, until we reach 8 turns dialogues.

\subsection{Reviewing phase and guidelines}
The data generated with the LMs include both HS and CN, therefore the annotators are allowed to post-edit both, unlike the previous session. The reviewing guidelines are similar to those for Session 1, with the following changes:

\begin{enumerate}[leftmargin=*]
    \item it is possible to swap single turns and not only pairs, since the connection between HS/CN is not granted a-priori as in the previous sessions\footnote{For example, a model can introduce hateful content when it is supposed to generate a CN, or viceversa (as showed in Table \ref{tab:session3_example1}, Appendix \ref{reviewing_examples}).}.
    \item if some turns in a dialogue have a clearly different target than the labeled one, they should try to change turns wording to fit the original target. 
    \item the annotators should check the veracity of fact-based statements since they might derive from LM hallucinations.
\end{enumerate}

\begin{table*}[htbp!]
\begin{center}\resizebox{\textwidth}{!}{
\begin{tabular}{l|rrr|rrcc|rrrrr} 
\hline
          & \multicolumn{3}{c|}{\textbf{Efficiency}}            & \multicolumn{4}{c}{\textbf{Quality}}    
          & \multicolumn{4}{|c}{\textbf{Syntactic Complexity}}  \\ 
\hline
          & \multicolumn{1}{c}{\textbf{del turns}} & \multicolumn{1}{c}{\textbf{HTER}} & \multicolumn{1}{c|}{\textbf{swap}} & \multicolumn{1}{c}{\textbf{RR$_{gen}$}} & \multicolumn{1}{c}{\textbf{RR$_{ed}$}} & \multicolumn{1}{c}{\textbf{NOV$_{g\text{-}g}$}} & \multicolumn{1}{c}{\textbf{NOV$_{g\text{-}e}$}} &
          \multicolumn{1}{|c}{\textbf{avg turn len}} &
          \textbf{avg turn \#} &
          \textbf{MSD} &
          \textbf{ASD} &
          \textbf{NST} 
          \\
Gold             & -                                                       & -                                                  & -                                                   & -                                                        & -                                                       & \multicolumn{1}{r}{-}                 & \multicolumn{1}{r|}{-}               & \underline{\textbf{25.873}}                                                     & 5.105                                                    & \underline{\textbf{5.515}}                                             & \underline{\textbf{4.722}}                                             & \underline{\textbf{1.785}}                                              \\
Session 1        & \underline{\textbf{12.381}}                                                  & \underline{\textbf{0.175}}                                              & \textbf{17.753}                                              & \textbf{9.112}                                                    & \textbf{5.382}                                                   & \multicolumn{1}{r}{\underline{\textbf{0.824}}}             & \multicolumn{1}{r|}{\underline{\textbf{0.821}}}           & \textbf{20.065}                                                     & \textbf{5.833}                                                    & \textbf{4.828}                                             & \textbf{4.236}                                             & 1.629                                              \\
Session 2        & -                                                       & \textbf{0.360}                                              & -                                                   & \underline{\textbf{4.244}}                                                    & \underline{\textbf{3.611}}                                                   & \multicolumn{1}{r}{0.756}             & \multicolumn{1}{r|}{0.770}           & 19.6108                                                    & 5.705                                                    & 4.778                                             & \textbf{4.236}                                             & 1.578                                              \\
Session 3        & \textbf{40.801}                                                  & 0.448                                              & \underline{\textbf{2.07}}                                                & 10.646                                                   & 6.385                                                   & \multicolumn{1}{r}{\textbf{0.795}}             & \multicolumn{1}{r|}{\textbf{0.800}}           & 19.944                                                     & \underline{\textbf{6.172}}                                                    & 4.757                                             & 4.112                                             & \textbf{1.655}                                             
\end{tabular}}
\caption{Results of the data collected at each session.}
\label{tab:across_sessions}
\end{center}
\end{table*}

\subsection{Results}\label{ch:exper}

Results of this session, in terms of \textit{efficiency} and \textit{quality}, are reported in Table \ref{tab:session3_results}.
There are two major conclusions we can draw\footnote{While our main focus is on dataset creation, the results of this session offer also a form of simple benchmarking and some useful insights for the development of new models. In fact, the various metrics that we employed (post-editing, turn deletion, etc.) already provide a good indication of the LMs performance, especially for an open-ended scenario.}.

Firstly, adding the post-edited dialogues obtained concatenating PAIRS$_{gold}$ to the training data (DIALO$_{gold}$) strongly increases the efficiency.
In fact, these models require much less deletion from the annotators with respect to the baselines, reaching a lower HTER ($<=0.4$). Also, even if the dialogues generated with the baselines have a higher novelty with respect to the training data, they are also extremely repetitive in almost all cases. 

Secondly, as already shown by \citet{tekiroglu2022using}, autoregressive models are producing more varied and relevant content as compared to seq2seq models. In fact, even if DialoGPT requires more post-editing than T5 configurations (with comparatively higher HTER scores), its output dialogues require a lower number of deletion. This indicates that that the DialoGPT generation is suboptimal but rarely unsuitable, while this often is not achieved by T5. In particular, turns swaps are present only for the DialoGPT models. According to the annotators, this is explained by the characteristics of T5 dialogues, which are more stereotypical, vague but have a better structure (see Table \ref{tab:session3_example2} in Appendix \ref{reviewing_examples}). This is also confirmed by the quality results: T5 models generate content with similar novelty scores to DialoGPT, but they also tend to be more repetitive.

\section{Session comparison \& Data description}
Finally, Table \ref{tab:across_sessions} compares the results for each session over the main metrics of interest. We observe that concatenating pre-existing material that is already verified (i.e. HS/CN from PAIRS$_{gold}$) requires less effort than generating new data from scratch or paraphrasing gold material, as Session 1 reaches a lower HTER than both Session 2 and Session 3. On the other hand, in terms of the structure of the dialogues, Session 1 requires the highest effort as shown by the high swap rate. Meanwhile, Session 2 is the least repetitive, but also the least novel, providing dialogues with a good wording, even if this is not accompanied with a novel content. In general, all the sessions reach an HTER lower or equal to 0.4, and similar novelty scores with respect to the gold data. Therefore, in all cases it was possible to enhance the novelty of the initial seed dataset, with a reasonable post-editing effort.

Table \ref{tab:across_sessions} also shows a syntactic analysis of the data collected with each session, calculated at turn-level. The dialogues generated with the Language Models achieve the most balanced distribution in terms of number of turns\footnote{We collected dialogues with 4, 6 and 8 turns, so a perfect balance would be of 6 turns.}, at the cost of simpler turns, as shown by the low maximum syntactic depth (MSD) and average syntactic depth (ASD) reached by Session 3. Paraphrasing instead provides the shortest generations both in terms of average turns length and of number of sentences (NST).

By comparing the results of the different sessions, we can conclude that the choice of the preferable data collection strategy firstly depends on the available input data, e.\ g.\ we might not always have gold HS/CN pairs available or multiple turns dialogues. Secondly, depending on the desired output, if the priority is to obtain novel content, Session 2 strategies would be the least favorite. Also, the concatenation of existing pairs as in Session 1 is a more cautious approach than the generation of completely new dialogues through LMs. Thus, Session 1 strategies can be preferred for a more conservative approach, whereas Session 3 strategies are better suited for a more creative data collection that comes at the cost of higher human correction effort.

\begin{table}[ht!]
\small
\begin{center}{
\begin{tabular}{l|rr}
\hline
 & \multicolumn{1}{|c}{\textbf{Dialogues}} & \multicolumn{1}{c}{\textbf{Coverage}} \\ \hline
\texttt{JEWS}            & 468                                    & 15.30                                 \\
\texttt{LGBT+}          & 591                                    & 19.32                                 \\
\texttt{MIGRANTS}        & 534                                    & 17.46                                 \\
\texttt{MUSLIMS}         & 505                                    & 16.51                                 \\
\texttt{POC}             & 493                                    & 16.12                                 \\
\texttt{WOMEN}           & 462                                    & 15.10                                 \\
Other           & 6                                      & 0.20   \\
\hline 
Total             & 3059                                   & 100                                  
\end{tabular}}
\caption{The distribution of targets in the final dataset. `Other' indicates a few cases of  intersectional targets among the 6 given, e.g. \texttt{MUSLIMS/WOMEN}.}
\label{tab:targets-distribution}
\end{center}
\end{table}

As a last step, we performed a sanity check in which a senior NGO expert conducted a qualitative evaluation by reading a random sample of the post-edited dialogues from each session. Their feedback was positive, no critical issues were raised and all the dialogues were approved both in terms of produced CNs and of their overall structure/naturalness.
Our final dataset, DIALOCONAN, includes also the dialogues collected through the various training phases and exploratory studies of the annotators. Table \ref{tab:targets-distribution} shows the distribution of targets in terms of number and percentage of dialogues. The distribution is reasonably balanced, with the \texttt{LGBT+} target being the most represented. Overall, we collected 3059 dialogues for a total of 16625 turns.

\section{Conclusion}\label{Conclusion}
In this paper we have presented a hybrid approach for dialogue data collection in the realm of hate speech countering. These dialogues have been obtained starting from two expert-based seed datasets and then combining the intervention of human annotators over machine generated dialogues. We tested 19 different strategies for generation, focusing on two crucial aspects of dialogue, i.e. structure and wording. We analysed all these strategies in terms of efficiency of the procedure and quality of the data obtained. 
The result of this work is DIALOCONAN, the first dataset comprising over 3000 fictitious multi-turn dialogues between a hater and an NGO operator, covering 6 targets of hate.

\section*{Acknowledgements}
We are deeply thankful to Stop Hate UK and its volunteers for the help in writing the dialogues for the DIALO$_{gold}$ seed dataset and for sharing their expertise, fundamental to this work.

\section*{Limitations}
The datasets currently available for CN generation are mainly for the English language and this one is no exception. The problem is that getting in contact with NGO operators for other languages is not easily solvable. The alternatives, such as translating this dataset into other languages represent a suboptimal solution. In fact each language and country has (i) its own peculiar canards against minorities, (ii) even if HS can be ported across countries (e.g. ``\textit{Migrants steal our jobs.}''), the arguments to counter such HS may vary (e.g. different laws, different socioeconomic situations, different statistical data). When translating dialogues all these nuances get lost reducing the possible effectiveness and introducing possible unnatural answers making reference to the country for which the original CN was written. 

Although we tried to keep the overall quality of the final output as high as possible, since the dataset is created through a human-machine collaboration paradigm, it can still not be on par with the data that we can potentially obtain with niche sourcing the whole dataset to skilled NGO operators. Additionally, the number of turns in dialogues are strictly controlled and might not reflect the more natural number of turns that would have occurred under those circumstances.

As previously stated, scraping NGO operators real online intervention is not desirable (we need to protect their identity). Still, even when collecting DIALO$_{gold}$ we encountered some problems, i.\ e.\ even if the annotators were trained and used to the task, they told us that the simulation was really frustrating (e.g. ``repeating over and over again the same hateful content'').

\section*{Ethics Statement}
Counter Narrative generation task and corresponding datasets have been proposed as a contribution of scientific research to a more ethical world. However, even the best intentions in the minefield of online hate can still bring along certain risks of undesired impacts on data curators (i.e., expert/non-expert annotators), on researchers, and on society. Therefore, in this study, we took meticulous precautions in order to avoid such effects.

\paragraph{Annotation Guidelines:} As the most important stakeholders of this research, the annotators were constantly supported in terms of mental welfare. In particular, we put in practice a mitigation procedure similar to the one proposed by \citet{vidgen-etal-2019-challenges}, as described in Section \ref{methodology}.

\paragraph{Dataset.} Since the dataset is created from scratch via an expert-machine collaboration schema (rather than scraping the dialogues among individuals online) it does not pose any threat to personal privacy or individual rights. Additionally, we avoid to model inappropriate CNs (e.\ g.\ containing abusive language) that could be produced by scraping non-expert users in their online activity \citep{mathew2018analyzing}. 

\paragraph{Generation Task.}  We consider the generation task as an aid to boost the data collection in terms of time, quantity, and certain quality aspects. Therefore, the models we trained are not meant to be deployed as part of a live system. Moreover, our main focus is clearly on the counter narrative generation part and the corresponding CN quality/diversity. For this reason, and to limit possible misuses, in our dialogues we tried to keep the HS as simple and stereotypical as possible and we always left `the last word' to a CN turn. We encourage other researchers to conduct the generation tasks in a similar manner and for this reason the dialogue dataset will be made available for research purposes together with the code/models used to generate it.

\bibliography{anthology,custom}
\bibliographystyle{acl_natbib}

\clearpage
\appendix

\section{Appendix}

\subsection{Session 1: Algorithms details}
\label{algo_concatenation}
The matching procedures over HS/CN pairs we employed are slightly different according to whether we performed a similarity (algorithm \ref{alg:similarity_connection}) or a keywords connection (algorithm \ref{alg:keywords_connection}). The main difference is that for similarity metrics it was always possible to choose among the 10 most similar pairs to the one of our interest, while when concatenating through keywords we put in practice an exact matching of pairs containing the same 2 keywords.

\begin{algorithm}[h]
\DontPrintSemicolon

\While{\small{\textsf{nr\_turns}} != {\small \textsf{desired\_nr\_turns}}:}{
\For{each \textit{$HS_i$}, \textit{$CN_i$}}{
 \eIf{\small{\textsf{nr\_turns}} == 0}{
$HS_{to\_match}$, $CN_{to\_match}$ \textleftarrow $HS_i$, $CN_i$

}{$HS_{to\_match}$, $CN_{to\_match}$ \textleftarrow {\small \textsf{chained\_dialo[-2]}, \textsf{chained\_dialo[-1]} } \;} {
\For{each \textit{$HS_j$}, \textit{$CN_j$}}{
\If{HS-HS connection}{
{\small \textbf{compute similarity}} ($HS_{to\_match}$, $HS_j$)
}
\If{CN-HS connection}{ {\small \textbf{compute similarity}} ($CN_{to\_match}$, $HS_j$) \; }
}
{\small \textbf{randomly select 1 pair from the top-10 most similar to}} $HS_{to\_match}$, $CN_{to\_match}$
} \;
\vspace{0.2pt}
{{\small \textsf{nr\_turns+=1}}}\;
\vspace{0.2pt}
{{\small \textsf{chained\_dialo}} += $HS_{selected}$, $CN_{selected}$}\;
\caption{Connection through the similarity of either HS-HS or CN-HS.} 
\label{alg:similarity_connection}
}
}
\end{algorithm}

\begin{algorithm}[h]
\DontPrintSemicolon
\While{\small{\textsf{nr\_turns}} != {\small \textsf{desired\_nr\_turns}}:}{
\For{each \textit{$HS_i$}, \textit{$CN_i$}}{
 \eIf{\small{\textsf{nr\_turns}} == 0}{
$HS_{to\_match}$, $CN_{to\_match}$ \textleftarrow $HS_i$, $CN_i$

}{$HS_{to\_match}$, $CN_{to\_match}$ \textleftarrow {\small \textsf{chained\_dialo[-2]}, \textsf{chained\_dialo[-1]} } \;} {
\For{each \textit{$HS_j$}, \textit{$CN_j$}}{
\If{HS-HS connection}{
{\small \textbf{find matching keywords}} ($HS_{to\_match}$, $HS_j$)
}
\If{CN-HS connection}{ {\small \textbf{find matching keywords}} ($CN_{to\_match}$, $HS_j$) \; }
}
{\small \textbf{randomly select 1 pair from those matching with}} $HS_{to\_match}$, $CN_{to\_match}$
} \;
\vspace{0.2pt}
{{\small \textsf{nr\_turns+=1}}}\;
\vspace{0.2pt}
{{\small \textsf{chained\_dialo}} += $HS_{selected}$, $CN_{selected}$}\;
\caption{Connection through HS-HS or CN-HS keywords matching.} 
\label{alg:keywords_connection}
}
}
\end{algorithm}

\subsection{Session 2: exploratory studies} 
\paragraph{Exploratory study 1} \label{exploratory_paraphrases}
We select three settings for paraphrasing with no style transfer.
We employ two paraphrasers: the Protaugment paraphraser and the Style transfer paraphraser. The tested configurations are the following:
\begin{itemize}
    \item Setting 1: Protaugment paraphraser with default parameters but \texttt{drop\_chance} is set to $0.1$ and \texttt{lower\_is\_better=False}.
    \item Setting 2: Protaugment paraphraser with default parameters but \texttt{lower\_is\_better=False}.
    \item Setting 3: Style transfer paraphraser with basic stye and $p=0.6$.
\end{itemize}

In total, we select 36 dialogues to be paraphrased: 12 for each setting, with 4 dialogues for 4, 6, and 8-turns dialogues. We generate 3 candidate paraphrases for each CN while the HS is not paraphrased, since our interest is to enlarge the CN data, and not the HS data. One expert annotator is given instructions of reading all the dialogues and, for each CN, to select the most appropriate paraphrasis and modify it to make it fit in the dialogue. The chosen paraphrasis should be the one which requires at the same time the least editing to fit in the dialogue naturally and to be as much different as possible from the original CN.

\begin{table*}[htbp]
\begin{center}\resizebox{\textwidth}{!}{
\begin{tabular}{l|rrr|rrr|rr}
\hline
 & \multicolumn{3}{c|}{\textbf{HTER}} &
\multicolumn{3}{c|}{\textbf{avg turn len}} &
\multicolumn{2}{c}{\textbf{$\Delta$ len}} 
\\
\hline
\multicolumn{1}{c}{\textbf{}} & \multicolumn{1}{|c}{\textbf{CN-$p_{sel}$}} & \multicolumn{1}{c}{\textbf{$p_{sel}$-$p_{ed}$}} & \multicolumn{1}{c|}{\textbf{CN-$p_{ed}$}} & \multicolumn{1}{c}{\textbf{CN}} & \multicolumn{1}{c}{\textbf{$p_{sel}$}} & \multicolumn{1}{c|}{\textbf{$p_{ed}$}} & \multicolumn{1}{c}{\textbf{CN-$p_{sel}$}} & \multicolumn{1}{c}{\textbf{CN-$p_{ed}$}} \\

Setting 1   & 0.55    & 0.46       & 0.61   & 24.67          & 21.83              & 21.44             & 11.51               & 13.09              \\
Setting 2   & 1.30    & 0.77       & 0.94   & 29.97          & 22.22              & 24.33             & 25.86               & 18.82              \\
Setting 3  & 0.85    & 0.44       & 0.78   & 26.17          & 22.17              & 23.58             & 15.28               & 9.90              
\end{tabular}}
\caption{Results of exploratory study 1: the metrics are calculated on CN only. CN is the original CN that was paraphrased, $p_{sel}$ the selected paraphrasis to be post-edited and $p_{ed}$ the post-edited paraphrasis.}
\label{tab:exp1_para_res}
\end{center}
\end{table*}

We aim for:
\begin{itemize}
    \item high values for the HTER between CN and original paraphrasis (HTER CN-$p_{sel}$) and between the CN and post-edited paraphrasis (HTER CN-$p_{ed}$);
    \item low HTER between original and post-edited paraphrasis (HTER $p_{sel}$-$p_{ed}$).
\end{itemize}

From the results in Table \ref{tab:exp1_para_res} we can notice that the first setting (Protaugment paraphreser with default setting but \texttt{lower\_is\_better = False} and \texttt{drop\_chance = 0.1}) is achieving the lowest values on the HTER between CN and $p_{sel}$ and between CN and $p_{ed}$, while the second lowest with the HTER between $p_{sel}$ and $p_{ed}$. The second setting (Protaugment paraphreser with default setting but \texttt{lower\_is\_better = False}) has the highest values on all the HTER results. The third setting (Style transfer paraphraser with default settings and basic style) has medium values on the HTER between CN and $p_{sel}$ and between CN and $p_{ed}$, but the lowest value on the HTER between $p_{sel}$ and $p_{ed}$, thus representing a good compromise for the characteristics of our interest. All the paraphrasers are making the original text shorter. From the results of the $\Delta$ length between CN and $p_{ed}$ we can notice that the setting 3 is the one that after post-editing is making the paraphrasis closer to the original length, whereas this is more difficult to achieve with the other settings (same effort, paraphrasis closer to the original CN length).

\begin{table}[htbp]{\begin{center}
\begin{tabular}{l|rrr}
\hline
 & \multicolumn{3}{|c}{\textbf{HTER}} 
 \\
 \hline
\multicolumn{1}{r}{\textbf{}} & \multicolumn{1}{|r}{\textbf{CN-$p_{sel}$}} & \multicolumn{1}{r}{\textbf{$p_{sel}$-$p_{ed}$}} & \multicolumn{1}{r}{\textbf{CN-$p_{ed}$}} \\
Setting 1            & 44.44      & 55.56            & 72.22     \\
Setting 2            & 100.00     & 86.11            & 100.00    \\
Setting 3            & 91.67      & 61.11            & 94.44    
\end{tabular}
\end{center}
}
\caption{The percentage of examples for each setting of exploratory study 1 with the HTER above the threshold value of 0.4. Results are calculated on CN only.}
\label{tab:exp1_para_res2}
\end{table}

As shown in Table \ref{tab:exp1_para_res2}, setting 1 is the one with less extreme results but for the HTER between CN-$p_{sel}$ and between $p_{sel}$-$p_{ed}$ the situation is the opposite than the one we aim for; setting 2 achieves the most extreme results. Despite setting 3 has a high percentage of examples reaching a high HTER between CN-$p_{sel}$ and between CN-$p_{ed}$, still the results for HTER between $p_{sel}$-$p_{ed}$ are not the worst. \\

For all these reasons, we decide to employ both the settings 1 and 3, while leaving out the setting 2.

\paragraph{Exploratory study 2}
In order to test the paraphrasis with style transfer, we use the following configurations:
\begin{itemize}
    \item Setting 1: Style former from casual to formal.
    \item Setting 2: Style former from formal to casual.
    \item Setting 3: Style transfer with Tweets style (split + 1 step pipeline)
    \item Setting 4: Style transfer with Tweets style (split + 2-steps pipeline)
    \item Setting 5: Style transfer with Switchboard style (no split + 1 step pipeline)
    \item Setting 6: Style transfer with Switchboard style (split + 1 step pipeline)
\end{itemize}

Once again, we select 12 dialogues for each setting, with 3 candidate paraphrases generated for each CN. The instructions given to the expert annotator are the same as in the first exploratory study.

\begin{table}[htbp]{\begin{center}
\begin{tabular}{c|rrr}
\hline
                               & \multicolumn{3}{c}{\textbf{HTER}}                                                                                   \\ \hline
\multicolumn{1}{r|}{\textbf{}} & \multicolumn{1}{c}{\textbf{CN-$p_{sel}$}} & \multicolumn{1}{c}{\textbf{$p_{sel}$-$p_{ed}$}} & \multicolumn{1}{c}{\textbf{CN-$p_{ed}$}} \\
Setting 1                      & 0.49                                 & 0.20                                 & 0.51                                  \\
Setting 2                      & 0.51                                 & 0.34                                 & 0.53                                  \\
Setting 3                      & 0.46                                 & 0.41                                 & 0.50                                  \\
Setting 4                      & 1.02                                 & 0.43                                 & 0.83                                  \\
Setting 5                      & 0.44                                 & 0.46                                 & 0.48                                  \\
Setting 5                      & 0.46                                 & 0.56                                 & 0.57                                 
\end{tabular}
\end{center}
}
\caption{HTER scores for the exploratory study 2. Results are calculated on CN only.}
\label{tab:exp2_para_res}
\end{table}

Results are showed in Table \ref{tab:exp2_para_res} and can be summed up as follows:
\begin{itemize}
    \item \textbf{Tweets}: setting 4 is achieving the highest HTER CN-$p_{sel}$ and HTER CN-$p_{ed}$ while having a HTER $p_{sel}$-$p_{ed}$ in the middle. We would prefer it to setting 3 which instead has almost the same HTER $p_{sel}$-$p_{ed}$ but a much lower HTER CN-$p_{sel}$ and HTER CN-$p_{ed}$;
    \item \textbf{Formal and informal}: both setting 1 and setting 2 achieve high HTER CN-$p_{sel}$ and HTER CN-$p_{ed}$ while low HTER $p_{sel}$-$p_{ed}$ with formal performing slightly better;
    \item \textbf{Switchboard}: setting 5 is preferable to setting 6 since it has a lower HTER $p_{sel}$-$p_{ed}$.
\end{itemize}
According to these results, we choose to employ setting 1, 2, 4 and 5 for the paraphrasis session.

\section{Session 3: Training details} \label{training_details}
For reproducibility purposes, we report here the parameters employed for fine-tuning the LMs used in Session 3. For each model, we used a version smaller than the largest available, i.\ e.\ the \textit{medium} version for DialoGPT and the \textit{base} version of T5. We used Optuna to conduct a hyperparameters search with 10 trials, and we selected the trial achieving the lowest evaluation loss. The search space for the parameters of our interest was the following: learning-rate: $\{1e-5,\ 2e-5,\ 3e-5,\ 4e-5,\ 5e-5\}$, warmup-ratio: $\{0,\ 0.1\}$, batch size: $\{1,\ 2,\ 4\}$, number of epochs: $\{2,\ 3,\ 5\}$. The selected parameters for each model are summed up in Table \ref{tab:training_details}.

\begin{table}[htbp]{\begin{center}
\begin{tabular}{l|rrrrr}
\hline
                                                                  & \multicolumn{1}{c}{\textbf{LR}} & \multicolumn{1}{l}{\textbf{BS}} & \multicolumn{1}{l}{\textbf{EP}} & \multicolumn{1}{c}{\textbf{WR}} & \multicolumn{1}{l}{\textbf{EL}} \\ \hline
DGPT$_b$                                                          & 3e-05                           & 2                               & 5                               & 0                               & 2.41                            \\
DGPT$_{mt}$                                                       & 4e-05                           & 2                               & 3                               & 0.1                             & 1.92                            \\
T5$_{b\text{-}1m}$                                                       & 4e-05                           & 4                               & 5                               & 0                               & 2.93                            \\
T5$_{mt\text{-}1m}$                                                      & 4e-05                           & 4                               & 3                               & 0                               & 2.26                            \\
\begin{tabular}[c]{@{}l@{}}T5$_{b\text{-}2m}$\\ (HS model)\end{tabular}  & 3e-05                           & 2                               & 5                               & 0.1                             & 3.12                            \\
\begin{tabular}[c]{@{}l@{}}T5$_{b\text{-}2m}$\\ (CN model)\end{tabular}  & 5e-05                           & 2                               & 2                               & 0.1                             & 2.86                            \\
\begin{tabular}[c]{@{}l@{}}T5$_{mt\text{-}2m}$\\ (HS model)\end{tabular} & 5e-05                           & 2                               & 2                               & 0.1                             & 2.41                            \\
\begin{tabular}[c]{@{}l@{}}T5$_{mt\text{-}2m}$\\ (CN model)\end{tabular} & 2e-05                           & 2                               & 2                               & 0.1                             & 2.32                           
\end{tabular}
\end{center}}
\caption{Training details for the models employed in Session 3.}
\label{tab:training_details}
\end{table}

\section{Reviewing examples} \label{reviewing_examples}

\begin{table*}[htbp]
\begin{center}\resizebox{\textwidth}{!}{
\begin{tabular}{l|l} 
\hline
\textbf{text}                                                                                                                                                                                                                                                                                & \multicolumn{1}{l}{\textbf{text$_{ed}$}}                                                                                                                                                                                                                                                       \\ 
\hline
\vcell{\begin{tabular}[b]{@{}l@{}}\textbf{HS$_0$: }Jews are only interested in themselves and their\\ own race.\end{tabular}}                                                                                                                                                                & \vcell{\colorbox{pink}{\textbf{HS$_2$:}} I hate Jews. They're just a bunch of hypocrites.}                                                                                                                                                                                             \\[-\rowheight]
\printcelltop                                                                                                                                                                                                                                                                                & \printcelltop                                                                                                                                \\
                                                                                                                                                                                                                                                                                               &                                                                                                                                                                                                                                                  \\
\vcell{\begin{tabular}[b]{@{}l@{}}\textbf{CN$_0$:} Do you really think that Jews only care about \\ themselves? Then why do they prefer to live in \\ our country where they are not the majority?\\ How would you answer that?\end{tabular}}                                                & \vcell{\begin{tabular}[b]{@{}l@{}}\colorbox{pink}{\textbf{CN$_2$:}} I'm not sure why you think that way. Why are all \\ Jews hypocrites?\end{tabular}}                                                                                                                                                       \\[-\rowheight]
\printcelltop                                                                                                                                                                                                                                                                                & \printcelltop                                                                                                                                  \\
                                                                                                                                                                                                                                                                                               &                                                                                                                                                                                                                                                \\
\vcell{\begin{tabular}[b]{@{}l@{}}\colorbox{lime}{\textbf{HS$_1$:}} Why are jews allowed to vote? They’re a \\ different religion and they shouldn't be allowed \\ to have their say.\end{tabular}}                                                                                                           & \vcell{\begin{tabular}[b]{@{}l@{}}\textbf{HS$_0$:} Jews are only interested in themselves and their \\ own race, but they still try to get into our society.\end{tabular}}                                                                                                                  \\[-\rowheight]
\printcelltop                                                                                                                                                                                                                                                                                & \printcelltop                                                                                                                               \\
                                                                                                                                                                                                                                                                                               &                                                                                                                                                                                                                                                   \\
\vcell{\begin{tabular}[b]{@{}l@{}}\colorbox{lime}{\textbf{CN$_1$:}} You seem to be misinterpreting the meaning of \\ the word “citizen”. The idea is that the individual is\\ allowed to choose their own religion just like \\ choosing how to vote, and that is the right of all \\ citizens.\end{tabular}} & \vcell{\begin{tabular}[b]{@{}l@{}}\textbf{CN$_0$:} Do you really think that Jews only care about \\ themselves? Then why do they prefer to live in our \\ country where they are not the majority? How would\\ you answer that?\end{tabular}}                                               \\[-\rowheight]
\printcelltop                                                                                                                                                                                                                                                                                & \printcelltop                                                                                                                                                                             \\
                                                                                                                                                                                                                                                                                               &                                                                                                                                                                                                     \\
\vcell{\colorbox{pink}{\textbf{HS$_2$:}} I hate Jews. They're just a bunch of hypocrites.}                                                                                                                                                                                                                    & \vcell{\begin{tabular}[b]{@{}l@{}}\colorbox{lime}{\textbf{HS$_1$:}} \textit{It's because they're hypocrites as I said!} They \\ shouldn't vote, they’re a different religion and they\\ shouldn't be allowed to have their say in our country.\end{tabular}}                                                          \\[-\rowheight]
\printcelltop                                                                                                                                                                                                                                                                                & \printcelltop                                                                                                                                  \\
                                                                                                                                                                                                                                                                                               &                                                                                                                                                                                                                                                \\
\vcell{\begin{tabular}[b]{@{}l@{}}\colorbox{pink}{\textbf{CN$_2$:}} I'm not sure why you think that way. Why \\ are all Jews hypocrites?\end{tabular}}                                                                                                                                                        & \vcell{\begin{tabular}[b]{@{}l@{}}\colorbox{lime}{\textbf{CN$_1$:}} You seem to be misinterpreting the meaning of \\ the word “citizen”. The idea is that the individual is \\ allowed to choose their own religion just like choosing \\ how to vote, and that is the right of all citizens.\end{tabular}}  \\[-\rowheight]
\printcelltop                                                                                                                                                                                                                                                                                & \printcelltop                                                                                                                                                                                                                                                                              
\end{tabular}
}
\caption{An example from Session 1 showing a turns swap and the modification of HS$_1$ to achieve linguistic alignment.}
\label{tab:session1_example1}
\end{center}
\end{table*}

Table \ref{tab:session1_example1} shows an example of turns swap from Session 1: CN$_2$ is a question that can be answered with HS$_0$, so it is moved at the beginning. At the same time, concluding the dialogue with the most substantial CN, i.\ e.\ CN$_1$, makes the dialogue stronger. HS$_1$ is modified by the addition of `because' in order to be linguistically aligned with the preceding turn, which is a question.

\begin{table*}[htbp]
\begin{center}\resizebox{\textwidth}{!}{
\begin{tabular}{l|l} 
\hline
\textbf{text}                                                                                                                                                                                              & \textbf{text$_{ed}$}                                                                                                                                                                                        \\ 
\hline
\textbf{HS$_0$: }Jews are genetically evil.                                                                                                                                                                & \textbf{HS$_0$: }Jews are genetically evil.                                                                                                                                                                 \\
                                                                                                                                                                                                          &                                                                                                                                                                                                             \\
\begin{tabular}[c]{@{}l@{}}\textbf{CN$_0$:} Just because they are Jewish, it does not\\mean they are bad people.\end{tabular}                                                                              & \begin{tabular}[c]{@{}l@{}}\textbf{CN$_0$:} Just because they are Jewish, it does not\\mean they are bad people.\end{tabular}                                                                               \\
                                                                                                                                                                                                          &                                                                                                                                                                                                             \\
\textbf{HS$_1$:} Jews are genetically evil.                                                                                                                                                                & \textbf{HS$_1$:} \textit{I have never met a Jew that wasn't evil, though.}                                                                                                                                      \\
                                                                                                                                                                                                          &                                                                                                                                                                                                             \\
\begin{tabular}[c]{@{}l@{}}\textbf{CN$_1$:} Not all Jews are bad, just an unlucky bunch \\who makes up a tiny part of the world's population, and\\this applies for every group in the world.\end{tabular} & \begin{tabular}[c]{@{}l@{}}\textbf{CN$_1$:} Not all Jews are bad, just an unlucky bunch \\who makes up a tiny part of the world's population, and\\this applies for every group in the world.\end{tabular}  \\
                                                                                                                                                                                                          &                                                                                                                                                                                                             \\
\colorbox{red}{\textbf{HS$_2$:}} Jews have a monopoly on evil.                                                                                                                                                             &                                                                                                                                                                                                             \\
                                                                                                                                                                                                          &                                                                                                                                                                                                             \\
\begin{tabular}[c]{@{}l@{}}\colorbox{red}{\textbf{CN$_2$:}} That's not true. And why would it be true? \\Why are they usually associated with evil?\end{tabular}                                                           &                                                                                                                                                                                                            
\end{tabular}
}
\caption{Example from Session 1 showing a particularly repetitive dialogue, for which it is necessary to delete two turns (HS$_2$, CN$_2$) and to modify HS$_1$.}
\label{tab:session1_example3}
\end{center}
\end{table*}

In table \ref{tab:session1_example3} an example of a dialogue resulting from the concatenation of similar HS is showed. The high repetitiveness makes it necessary to remove the pair HS$_2$, CN$_2$ and to modify HS$_1$.

\begin{table*}[htbp]
\begin{center}\resizebox{\textwidth}{!}{
\begin{tabular}{p{0.30\linewidth} |p{0.3\linewidth} |p{0.3\linewidth}} 
\hline
\textbf{text}                                                                                                                                                                                                 & \textbf{$p_{sel}$}                                                                                                                             & \textbf{$p_{ed}$}                                                                                                                                                                                              \\ 
\hline
\vspace{0.001pt}
\vcell{\textbf{HS$_0$: }black folks are very dangerous.}                                                                                                                                                      & \vcell{}                                                                                                                                        & \vcell{\textbf{HS$_0$: }black folks are very dangerous.}                                                                                                                                                        \\[-\rowheight]
\printcelltop                                                                                                                                                                                                 & \printcellmiddle                                                                                                                                & \printcelltop                                                                                                                                                                                                   \\
                                                                                                                                                                                                              &                                                                                                                                                 &                                                                                                                                                                                                                 \\
\vcell{\textbf{CN$_0$: }Why do you say that? Do you have any sources for this? I would appreciate it if you would provide such a source.}                                                                     & \vcell{\textbf{CN$_0$: }Why is this even a thing. Is there any source for this. If you could provide such source, I would appreciate it.}       & \vcell{\textbf{CN$_0$: }\textit{Is this even a thing? Is there any source for this?} If you could provide such source, I would appreciate it.}                                                                           \\[-\rowheight]
\printcelltop                                                                                                                                                                                                 & \printcellmiddle                                                                                                                                & \printcelltop                                                                                                                                                                                                   \\
                                                                                                                                                                                                              &                                                                                                                                                 &                                                                                                                                                                                                                 \\
\vcell{\textbf{HS$_1$: }Blacks are criminals, that's just who they are. They are a threat to society.}                                                                                                        & \vcell{}                                                                                                                                        & \vcell{\textbf{HS$_1$: }Blacks are criminals, that's just who they are. They are a threat to society.}                                                                                                          \\[-\rowheight]
\printcelltop                                                                                                                                                                                                 & \printcellmiddle                                                                                                                                & \printcelltop                                                                                                                                                                                                   \\
                                                                                                                                                                                                              &                                                                                                                                                 &                                                                                                                                                                                                                 \\
\vcell{\textbf{CN$_1$: }Such statements are just fear-mongering. There is no such thing as a criminal race.}                                                                                                  & \vcell{\textbf{CN$_1$: }Such statements are just fear mongering There is no such thing as a criminal race.}                                     & \vcell{\textbf{CN$_1$: }\textit{This is a fearmongering statement. Criminal races don't exist.}}                                                                                                                         \\[-\rowheight]
\printcelltop                                                                                                                                                                                                 & \printcellmiddle                                                                                                                                & \printcelltop                                                                                                                                                                                                   \\
                                                                                                                                                                                                              &                                                                                                                                                 &                                                                                                                                                                                                                 \\
\vcell{\textbf{HS$_2$: }Really? Because blacks are a plague on society. They are vicious and cruel.They steal corrupt the criminal justice system. They are basically just a bunch of thieves and murderers.} & \vcell{}                                                                                                                                        & \vcell{\textbf{HS$_2$: }Really? Because blacks are a plague on society. They are vicious and cruel. They steal corrupt the criminal justice system. They are basically just a bunch of thieves and murderers.}  \\[-\rowheight]
\printcelltop                                                                                                                                                                                                 & \printcellmiddle                                                                                                                                & \printcelltop                                                                                                                                                                                                   \\
                                                                                                                                                                                                              &                                                                                                                                                 &                                                                                                                                                                                                                 \\
\vcell{\textbf{CN$_2$: }Criminality and corruption are not the province of any specific racial or culture group.}                                                                                             & \vcell{\textbf{CN$_2$: }Criminality and corruption are not a province of any specific racial or culture group.}                                 & \vcell{\textbf{CN$_2$: }Criminality and corruption \textit{can't be tied to any specific racial or culture group.}}                                                                                                      \\[-\rowheight]
\printcelltop                                                                                                                                                                                                 & \printcellmiddle                                                                                                                                & \printcelltop                                                                                                                                                                                                   \\
                                                                                                                                                                                                              &                                                                                                                                                 &                                                                                                                                                                                                                 \\
\vcell{\textbf{HS$_3$: }You still haven't convinced me that black people aren't the source of crime in our society.}                                                                                          & \vcell{}                                                                                                                                        & \vcell{\textbf{HS$_3$: }You still haven't convinced me that black people aren't the source of crime in our society.}                                                                                            \\[-\rowheight]
\printcelltop                                                                                                                                                                                                 & \printcellmiddle                                                                                                                                & \printcelltop                                                                                                                                                                                                   \\
                                                                                                                                                                                                              &                                                                                                                                                 &                                                                                                                                                                                                                 \\
\vcell{\textbf{CN$_3$: }It is not true that~ blacks are the source of crime in our society. The problem is violence and poverty, not race.}                                                                   & \vcell{\textbf{CN$_3$: }Not true blacks are the source of crime in our society We need to find the problem in violent and poverty, not racism.} & \vcell{\textbf{CN$_3$: }\textit{That blacks} are the source of crime in our society \textit{is simply not true}. We need to \textit{search} the problem in \textit{violence} and poverty, not \textit{race}.}                                                \\[-\rowheight]
\printcelltop                                                                                                                                                                                                 & \printcellmiddle                                                                                                                                & \printcelltop                                             \end{tabular}}
\caption{An example of post-editing of the selected paraphrases (para$_{sel}$) of the CN.}
\label{tab:session2_example1}
\end{center}
\end{table*}

In table \ref{tab:session2_example1} an example of CN post-editing coming from Session 2 is showed: the selected paraphrasis is modified in order to be as much different from the original as possible, while keeping the dialogue flow naturally. The paraphrases of CN$_1$ and CN$_2$ are highly similar to the original text, and require a major intervention from the annotator.

\begin{table*}[htbp]
\begin{center}\resizebox{\textwidth}{!}{
\begin{tabular}{p{0.50\linewidth} |p{0.5\linewidth}} 

\hline
\textbf{text}                                                                                                                                                & \textbf{text$_{ed}$}                                                                                                                                                                                  \\ 
\hline
\vspace{0.001pt}
\vcell{\textbf{HS$_0$: }The problem with Islam is that it is about killing infidels. It must be banned.}                                                     & \vcell{\textbf{HS$_0$: }The problem with Islam is that it is about killing infidels. It must be banned.}                                                                                            \\[-\rowheight]
\printcelltop                                                                                                                                                & \printcelltop                                                                                                                                                                                       \\
                                                                                                                                                             &                                                                                                                                                                                                     \\
\vcell{\textbf{CN$_0$: }Islam is about love, not death.}                                                                                                     & \vcell{\textbf{CN$_0$: }Islam is about love, not death.}                                                                                                                                            \\[-\rowheight]
\printcelltop                                                                                                                                                & \printcelltop                                                                                                                                                                                       \\
                                                                                                                                                             &                                                                                                                                                                                                     \\
\vcell{\textbf{HS$_1$: }It is also about conquest and murder. I can't understand why anybody would want to murder infidels.}                                 & \vcell{\textbf{HS$_1$: }It is also about conquest and murder. I can't understand why anybody would want to murder infidels.}                                                                        \\[-\rowheight]
\printcelltop                                                                                                                                                & \printcelltop                                                                                                                                                                                       \\
                                                                                                                                                             &                                                                                                                                                                                                     \\
\vcell{\textbf{CN$_1$: }This is about human dignity and respect for people's rights. To murder infidels would be to remove human dignity and human respect.} & \vcell{\textbf{CN$_1$: }To murder infidels would be to remove human dignity and human respect, and since Islam cares for human rights and human dignity, I can assure you that it doesn't happen.}  \\[-\rowheight]
\printcelltop                                                                                                                                                & \printcelltop                                                                                                                                                                                       \\
                                                                                                                                                             &                                                                                                                                                                                                     \\
\vcell{\textbf{HS$_2$: }Islam is an outdated and violent religion, which is incompatible with Western culture.}                                              & \vcell{\textbf{HS$_2$: }Are you sure? Islam is an outdated and violent religion, which is incompatible with Western culture.}                                                                       \\[-\rowheight]
\printcelltop                                                                                                                                                & \printcelltop                                                                                                                                                                                       \\
                                                                                                                                                             &                                                                                                                                                                                                     \\
\vcell{\textbf{CN$_2$: }It is not incompatible with Western culture, but Islam is a dead cult and should be eliminated.}                                     & \vcell{\textbf{CN$_2$: }It is not incompatible with Western culture, \textit{Islam has existed in Europe for centuries and is part of our culture.}}                                                         \\[-\rowheight]
\printcelltop                                                                                                                                                & \printcelltop                                                                                                                                                                                       \\
                                                                                                                                                             &                                                                                                                                                                                                     \\
\vcell{\colorbox{pink}{\textbf{HS$_3$:}} Islam is a religion, and it is not a dead cult. It is not only about killing people, but also about love and peace.}                 & \vcell{\colorbox{lime}{\textbf{CN$_3$:}} Islam is a totalitarian ideology that promotes rape, torture, and murder. I don't want it in my country!}                                                                   \\[-\rowheight]
\printcelltop                                                                                                                                                & \printcelltop                                                                                                                                                                                       \\
                                                                                                                                                             &                                                                                                                                                                                                     \\
\vcell{\colorbox{lime}{\textbf{CN$_3$:}} Islam is a totalitarian ideology that promotes rape, torture, and murder.}                                                           & \vcell{\colorbox{pink}{\textbf{HS$_3$:}} Islam is a religion, not a totalitarian ideology. It is not about killing people, only about love and peace, so it has the same values as our country.}                     \\[-\rowheight]
\printcelltop                                                                                                                                                & \printcelltop                                                                                                                                                                                      
\end{tabular}}
\caption{Example from the annotation of Session 3.}
\label{tab:session3_example1}
\end{center}
\end{table*}

Table \ref{tab:session3_example1} and Table \ref{tab:session3_example2} show two peculiar cases of the annotators' intervention in Session 3. In table \ref{tab:session3_example1}, HS$_3$ and CN$_3$ are swapped because CN$_3$ contains hateful content , while HS$_3$ is a CN. For the same reason it was necessary to post-edit CN$_2$.

In table \ref{tab:session3_example2}, an example of a dialogue generated with T5, characterised by a poorly varied content. Both HS and CN are edited a lot to make the dialogue more natural.

\begin{table*}[htbp]
\begin{center}\resizebox{\textwidth}{!}{
\begin{tabular}{p{0.50\linewidth} |p{0.5\linewidth}} 
\hline
\textbf{text}                                                                                                                          & \textbf{text$_{ed}$}                                                                                                                                                  \\ 
\hline
\vspace{0.001pt}
\vcell{\textbf{HS$_0$: }Women are nothing but objects, unworthy of respect and protection.}                                            & \vcell{\textbf{HS$_0$: }Women are nothing but objects, unworthy of respect and protection.}                                                                           \\[-\rowheight]
\printcelltop                                                                                                                          & \printcelltop                                                                                                                                                         \\
\vcell{}                                                                                                                               & \vcell{}                                                                                                                                                              \\[-\rowheight]
\printcelltop                                                                                                                          & \printcelltop                                                                                                                                                         \\
\vcell{\textbf{CN$_0$:} Women are human beings, they deserve respect and protection.}                                                  & \vcell{\textbf{CN$_0$:} Women are human beings, \textit{not objects, and they're absolutely worthy of respect and protection, like all human beings}.}                         \\[-\rowheight]
\printcelltop                                                                                                                          & \printcelltop                                                                                                                                                         \\
\vcell{}                                                                                                                               & \vcell{}                                                                                                                                                              \\[-\rowheight]
\printcelltop                                                                                                                          & \printcelltop                                                                                                                                                         \\
\vcell{\textbf{HS$_1$: }Women are animals too, they need to be protected.}                                                             & \vcell{\textbf{HS$_1$:} \textit{Also animals need to be protected sometimes, so women must be animals}.}                                                                       \\[-\rowheight]
\printcelltop                                                                                                                          & \printcelltop                                                                                                                                                         \\
\vcell{}                                                                                                                               & \vcell{}                                                                                                                                                              \\[-\rowheight]
\printcelltop                                                                                                                          & \printcelltop                                                                                                                                                         \\
\vcell{\textbf{CN$_1$:} Women are not animals, they just have rights and should be treated as such.}                                   & \vcell{\textbf{CN$_1$:} \textit{Animals rights exist, but women are protected under human rights because they're humans and not animals. That's the difference.}}              \\[-\rowheight]
\printcelltop                                                                                                                          & \printcelltop                                                                                                                                                         \\
\vcell{}                                                                                                                               & \vcell{}                                                                                                                                                              \\[-\rowheight]
\printcelltop                                                                                                                          & \printcelltop                                                                                                                                                         \\
\vcell{\textbf{HS$_2$:} They're nothing but objects to be exploited for.}                                                              & \vcell{\textbf{HS$_2$: }No, they're nothing but objects to be exploited for.}                                                                                         \\[-\rowheight]
\printcelltop                                                                                                                          & \printcelltop                                                                                                                                                         \\
\vcell{}                                                                                                                               & \vcell{}                                                                                                                                                              \\[-\rowheight]
\printcelltop                                                                                                                          & \printcelltop                                                                                                                                                         \\
\vcell{\textbf{CN$_2$:} Women are human beings with human rights and that includes the right to life.}                                 & \vcell{\textbf{CN$_2$:} \textit{You should check the Universal Declaration of Human Rights, especially the section where it says that men and women have the same rights.}}    \\[-\rowheight]
\printcelltop                                                                                                                          & \printcelltop                                                                                                                                                         \\
\vcell{}                                                                                                                               & \vcell{}                                                                                                                                                              \\[-\rowheight]
\printcelltop                                                                                                                          & \printcelltop                                                                                                                                                         \\
\vcell{\textbf{HS$_3$: }Well, women don't deserve human rights, they are parasitic and steal our jobs.}                                & \vcell{\textbf{HS$_3$: }Well, women don't deserve human rights, they are parasitic and steal our jobs.}                                                               \\[-\rowheight]
\printcelltop                                                                                                                          & \printcelltop                                                                                                                                                         \\
\vcell{}                                                                                                                               & \vcell{}                                                                                                                                                              \\[-\rowheight]
\printcelltop                                                                                                                          & \printcelltop                                                                                                                                                         \\
\vcell{\textbf{CN$_3$: }Women often work hard for themselves, but they do not have to steal their jobs, it's because of their gender.} & \vcell{\textbf{CN$_3$: }Women work hard for themselves \textit{and their families}, they don't steal jobs \textit{but simply apply for the ones they want, like it's in their right.}}  \\[-\rowheight]
\printcelltop                                                                                                                          & \printcelltop                                                                                                                                                        
\end{tabular}}
\caption{Example from the annotation of Session 3.}
\label{tab:session3_example2}
\end{center}
\end{table*}

\end{document}